\documentclass{article} 
\usepackage{iclr2026_conference,times}

\usepackage{amsmath,amsfonts,bm}

\def\eqref#1{equation~\ref{#1}}

\def\1{\bm{1}}

\DeclareMathAlphabet{\mathsfit}{\encodingdefault}{\sfdefault}{m}{sl}
\SetMathAlphabet{\mathsfit}{bold}{\encodingdefault}{\sfdefault}{bx}{n}

\usepackage{hyperref}
\usepackage{url}
\usepackage{xcolor}
\usepackage{multirow}
\usepackage{booktabs}
\usepackage{makecell}
\usepackage{float} 
\usepackage{graphicx}
\usepackage[skip=5pt]{caption} 
\usepackage{fontawesome5}
\usepackage{xcolor}
\usepackage{twemojis}
\usepackage[T1]{fontenc}

\iclrfinalcopy 
\begin{document}

\begin{center}
    \vspace*{-3.5em}

    {\LARGE \textbf{GLM-OCR Technical Report}}

    \vspace{0.6em}

    {\normalsize \bfseries
    Shuaiqi Duan$^{\star 1}$ \quad
    Yadong Xue$^{\star 1}$ \quad
    Weihan Wang$^{\star 1}$ \quad
    Zhe Su$^1$ \quad
    Huan Liu$^1$ \quad \\
    Sheng Yang$^1$ \quad
    Guobing Gan$^1$ \quad
    Guo Wang$^1$ \quad
    Zihan Wang$^1$ \quad
    Shengdong Yan$^1$ \quad \\
    Dexin Jin$^1$ \quad
    Yuxuan Zhang$^1$ \quad
    Guohong Wen$^1$ \quad
    Yanfeng Wang$^1$ \quad 
    Yutao Zhang$^1$ \quad \\
    Xiaohan Zhang$^1$ \quad
    Wenyi Hong$^1$ \quad
    Yukuo Cen$^1$ \quad
    Da Yin$^1$ \quad
    Bin Chen$^1$ \quad \\
    Wenmeng Yu$^{\dag 1}$ \quad
    Xiaotao Gu$^{\dag 1}$ \quad
    Jie Tang$^2$
     }

    \vspace{0.4em}

{\normalsize
    $^1$Zhipu AI \quad $^2$Tsinghua University \\[2pt]
    $^\star$Equal contribution $\quad$ $^\dag$Project leader
}

    \vspace{0.5em}

    \scriptsize
    \renewcommand{\arraystretch}{1.0}
    \begin{tabular}{@{}l@{\hspace{2mm}}l}
        \faGithub\ \textbf{Code:} & \url{https://github.com/zai-org/GLM-OCR} \\
        \includegraphics[height=1.8ex]{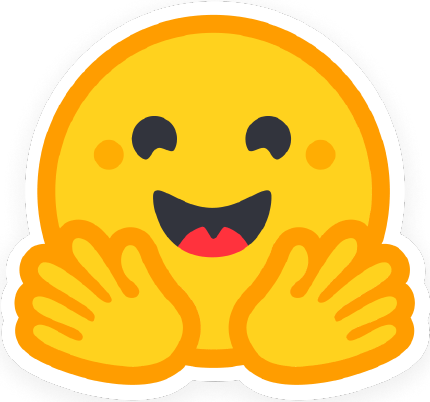} \textbf{Model:} & \url{https://huggingface.co/zai-org/GLM-OCR} \\
        \faMicrophone\ \textbf{Demo:} & \url{https://ocr.z.ai/} \\
    \end{tabular}
\end{center}

\begin{abstract}

GLM-OCR is an efficient 0.9B-parameter compact multimodal model designed for real-world document understanding. It combines a 0.4B-parameter CogViT visual encoder with a 0.5B-parameter GLM language decoder, achieving a strong balance between computational efficiency and recognition performance. To address the inefficiency of standard autoregressive decoding in deterministic OCR tasks, GLM-OCR introduces a Multi-Token Prediction (MTP) mechanism that predicts multiple tokens per step, significantly improving decoding throughput while keeping memory overhead low through shared parameters. At the system level, a two-stage pipeline is adopted: PP-DocLayout-V3 first performs layout analysis, followed by parallel region-level recognition. Extensive evaluations on public benchmarks and industrial scenarios show that GLM-OCR achieves competitive or state-of-the-art performance in document parsing, text and formula transcription, table structure recovery, and key information extraction. Its compact architecture and structured generation make it suitable for both resource-constrained edge deployment and large-scale production systems.

\end{abstract}

\begin{figure}[!h]
\centering
\includegraphics[width=\linewidth]{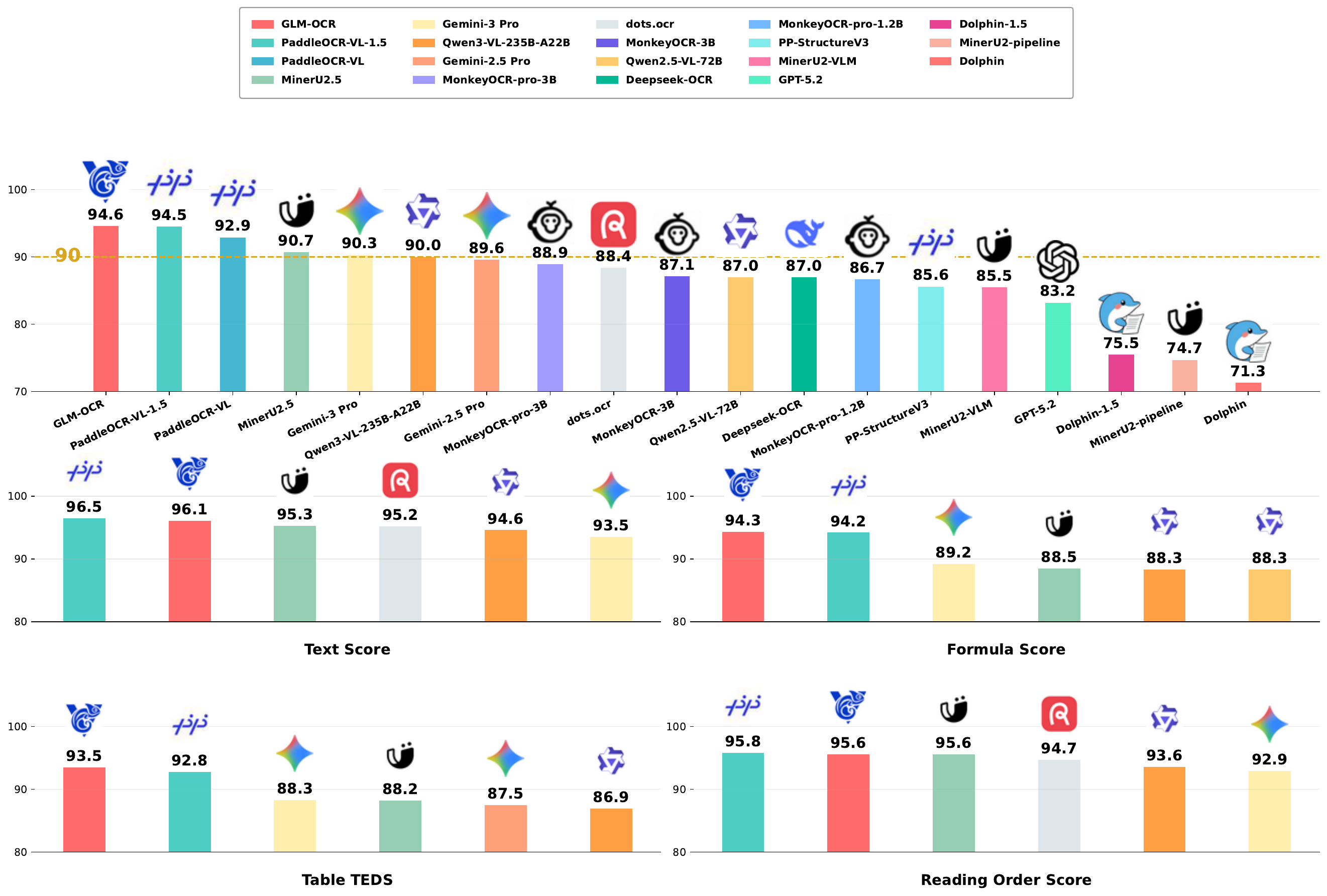}
\caption{Performance of GLM-OCR on OmniDocBench v1.5.}
\label{fig:eval_comparison}
\end{figure}

\newpage

\tableofcontents
\newpage

\section{Introduction}

Document understanding is a core capability in modern information systems, supporting the extraction and structuring of knowledge from visually rich and layout-intensive documents such as financial reports, scientific articles, contracts, and invoices. Traditional OCR systems~\cite{Tesseract,EasyOCR,PaddleOCR} mainly focus on plain text transcription and rely on multi-stage pipelines with handcrafted rules for layout parsing and downstream information extraction. While effective for simple scenarios, these approaches often struggle with complex layouts, diverse document formats, and real-world production requirements.

Recent multimodal large language models (MLLMs)~\cite{Qwen3-VL,glm-4.5v,seed1_5vl} unify visual perception and language understanding within a single framework and significantly improve document understanding performance. However, their large model size and autoregressive decoding paradigm lead to high computational cost, slow inference, and substantial memory consumption, which makes large-scale deployment under high-concurrency or edge environments challenging.

In practical production systems, document intelligence solutions must simultaneously provide: (1) strong performance on complex content such as tables, formulas, code, and seals, (2) high-throughput and low-latency inference, and (3) flexible integration and domain adaptation. GLM-OCR is developed to address these system-level requirements within a unified multimodal framework.

\textbf{GLM-OCR} is a lightweight multimodal OCR model for comprehensive document understanding. Built on the GLM-V encoder–decoder framework~\cite{glm-4.5v}, it combines a 0.4B-scale CogViT visual encoder trained in large-scale image–text data, a lightweight cross-modal connector and a 0.5B-scale GLM language decoder~\cite{glm2024chatglm}. The entire model contains only 0.9B parameters, enabling high-throughput and low-latency inference while maintaining strong recognition performance.

Beyond architectural optimization, GLM-OCR also considers the mismatch between conventional autoregressive generation and the characteristics of OCR tasks. OCR is inherently a deterministic task with strong local dependencies and explicit structural supervision, where strictly autoregressive token-by-token decoding is inefficient. Therefore, we introduce Multi-Token Prediction (MTP)~\cite{deepseekv3} into both training and inference. MTP enables the simultaneous prediction of multiple tokens, substantially improving training efficiency and decoding throughput while preserving recognition accuracy, and is particularly advantageous for long structured outputs such as tables. To control the additional memory overhead introduced by MTP, we further adopt a parameter-sharing scheme across the draft models, which substantially reduces the additional GPU memory overhead~\cite{glm5}. In practice, GLM-OCR is trained to predict ten tokens per step and generates 5.2 tokens per decoding step on average at inference time, bringing approximately 50\% throughput improvement.

At the system level, GLM-OCR adopts a two-stage pipeline consisting of layout analysis and parallel content recognition. The layout stage is powered by PP-DocLayout-V3~\cite{PaddleOCR-VL-1.5}, which detects structured regions and enables parallel recognition across different document areas. This design improves both robustness and processing efficiency for complex real-world documents.

\textbf{Results.} Figure~\ref{fig:eval_comparison} shows that GLM-OCR achieves 94.6 on OmniDocBench v1.5~\cite{OmniDocBench}, ranking first among all evaluated models despite its compact 0.9B size. Besides, GLM-OCR delivers strong performance across text recognition, formula recognition, table parsing, and key information extraction, reaching 94.0 on OCRBench (Text)~\cite{OCRBench} and 96.5 on UniMERNet~\cite{UniMERNet}, and achieving 85.2 on PubTabNet~\cite{PubTabNet} and 86.0 on TEDS~\footnote{https://modelscope.cn/datasets/jockerK/TEDS\_TEST}. It also performs competitively on information extraction benchmarks such as Nanonets-KIE and Handwritten-Forms~\cite{IDPLeaderboard}, with performance comparable to significantly larger general multimodal models.

In addition to public benchmarks, we evaluate GLM-OCR on six high-frequency real-world scenarios, including code document parsing, natural-scene table recognition, handwritten text recognition, multilingual OCR~\footnote{Chinese, English, French, Spanish, Russian, German, Japanese, and Korean}, seal recognition, and receipt KIE. GLM-OCR consistently delivers strong results across all settings, achieving 91.5 on real-world table recognition, 90.5 on seal recognition, and 94.5 on receipt KIE. These results indicate that GLM-OCR generalizes beyond curated benchmarks and remains effective under practical production conditions.

\textbf{Deployment and Finetuning.} GLM-OCR supports efficient inference with modern serving frameworks such as vLLM~\cite{vllm}~\footnote{https://github.com/vllm-project/vllm}, SGLang~\footnote{https://github.com/sgl-project/sglang}, and Ollama~\footnote{https://github.com/ollama/ollama}, enabling deployment in both large-scale and resource-constrained edge scenarios. It also provides full finetuning support through LLaMA-Factory~\footnote{https://github.com/hiyouga/LlamaFactory}, enabling rapid adaptation to domain-specific document understanding tasks~\footnote{Tutorial: https://github.com/zai-org/GLM-OCR/blob/main/examples/finetune/README.md}.

\section{Methodology}
\label{sec:methodology}

In this section, we present the overall design of the GLM-OCR framework, including its architectural components, task formulation, and training strategy. We first introduce the core design motivations that guide our system, followed by a detailed description of the model architecture and task-specific pipelines. Finally, we elaborate on the multi-stage training recipe that progressively aligns visual and language representations, enhances structured generation ability, and optimizes task performance through supervised and reinforcement learning.

\begin{figure}[!h]
\centering
\includegraphics[width=\linewidth]{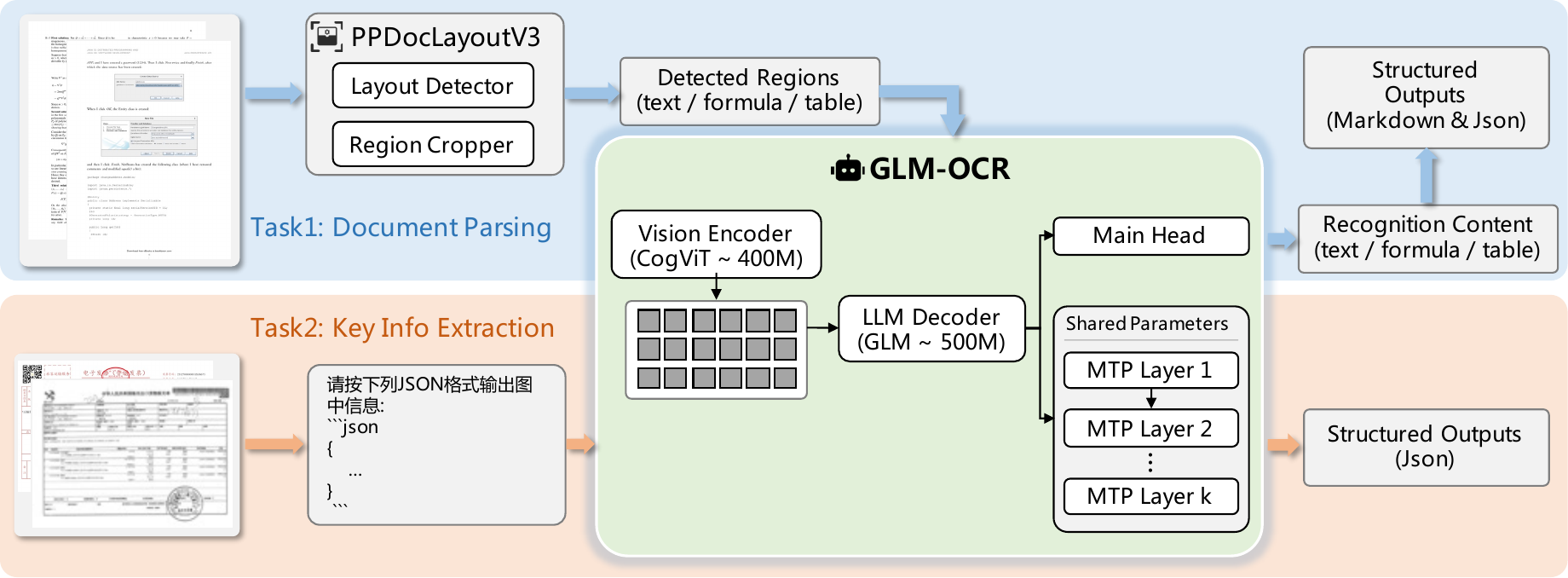}
\caption{Overall architecture and workflow of the GLM-OCR framework. This system supports two primary tasks: Document Parsing (Task 1), which combines layout detection and region cropping to produce structured Markdown and JSON outputs; and Key Information Extraction (Task 2), which directly extracts structured JSON data based on input visual prompts.}
\label{fig:eval_comparison}
\end{figure}

\subsection{Model Overview}

To provide a comprehensive understanding of the GLM-OCR architecture, we first discuss the primary design rationale behind our approach, followed by a detailed description of the model structure and its execution across different tasks.

\paragraph{Motivation.} Three fundamental observations and goals in document understanding inform our architectural design:

\textbf{1. Integration of Layout Analysis in Document Parsing:} 
In practice, we observe that small-scale models are highly susceptible to hallucinations and repetitive generation when processing documents with complex layouts. By explicitly introducing a layout analysis module prior to recognition, we decompose complex layout structures into multiple simpler sub-problems, significantly enhancing the overall performance and stability of the model. Furthermore, partitioning large, complex pages into smaller, independent regions allows for parallel recognition, which substantially improves inference efficiency.

\textbf{2. Incorporation of Key Information Extraction (KIE):} 
Document parsing and KIE can both be formulated as structured generation problems conditioned on visual inputs. While document parsing focuses on reconstructing the full structural representation of a document (e.g., Markdown or JSON), KIE aims to generate task-specific structured fields from the same visual source. From a modeling perspective, both tasks require:
(i) robust visual-text alignment,
(ii) structural reasoning over layout and semantic regions, and
(iii) the ability to generate well-formed structured outputs.

Therefore, instead of treating them as isolated pipelines, we unify them under a shared generative framework. This design encourages the model to learn generalizable document-level representations while leveraging task-specific prompts to control output formats. Such a unified formulation improves parameter efficiency and promotes cross-task knowledge transfer without introducing additional architectural complexity.

\textbf{3. Adoption of Multi-Token Prediction (MTP):} 
Standard LLM decoding generates one token at a time, which can be computationally expensive and slow for long-form document generation. We employ Multi-token Prediction (MTP) to address two primary challenges:
\begin{itemize}
    \item \textit{Inference Speed:} By predicting $k$ tokens simultaneously, we significantly reduce the total number of decoding steps.
    \item \textit{Contextual Modeling:} MTP encourages the model to plan further ahead. This is particularly beneficial in OCR tasks where structural tokens (e.g., table tags or Markdown syntax) exhibit strong local dependencies. Consequently, this approach yields fewer "broken" tags and produces more robust structured outputs.
\end{itemize}

\paragraph{Architecture.} 
As illustrated in Figure~\ref{fig:eval_comparison}, the system is centered around the GLM-OCR Core, which follows a vision-language generative paradigm. The core model consists of:

\begin{itemize}
    \item \textbf{Vision Encoder (CogViT, 400M parameters):} Responsible for extracting high-level visual representations from document images.
    \item \textbf{LLM Decoder (GLM, 500M parameters):} An autoregressive language model that generates structured textual outputs conditioned on visual embeddings and textual prompts.
\end{itemize}

The visual features produced by the encoder are projected into the language embedding space and fed into the decoder as prefix tokens. During decoding, the model predicts structured outputs (e.g., Markdown or JSON) in an autoregressive manner.

To improve decoding efficiency and structural consistency, we introduce a Multi-Token Prediction (MTP) mechanism. In addition to the main prediction head, we attach $k$ shared-parameter auxiliary heads that simultaneously predict the next $k$ tokens. These heads share the same parameters but are trained to model different future offsets. During inference, this allows the model to generate multiple tokens per step, reducing latency while encouraging better local structural coherence. 

The framework supports two primary tasks under a unified generative formulation:

\textbf{Task 1: Document Parsing.} 

Given a document image, the pipeline first performs layout analysis using PPDocLayoutV3, which decomposes the document into semantically coherent regions (e.g., paragraphs, tables, formulas). Each region is independently processed by the GLM-OCR Core.

The generated regional outputs are subsequently aggregated by a Merge \& Post Process module, which restores the reading order and produces structured outputs in Markdown and JSON formats. This modular design reduces hallucination risks, improves robustness to complex layouts, and enables parallel processing of document regions.

\textbf{Task 2: Key Information Extraction.} 

For KIE, the full document image is directly fed into the GLM-OCR Core together with a task-specific textual prompt (e.g., instructing the model to extract invoice fields in JSON format). Unlike document parsing, this task does not rely on explicit layout cropping. Instead, the model learns to attend to relevant visual regions implicitly under prompt guidance.

Both tasks are thus unified as conditional structured generation problems, differing only in preprocessing strategy and prompt specification.

\subsection{Training Recipe}

The training process of GLM-OCR is divided into several distinct stages, systematically progressing from vision-language alignment to task-specific refinement and reinforcement learning. The detailed training recipe, including the data types, learning rates, and training scale for each phase, is summarized in Table \ref{tab:training_recipe}.

\begin{table}[htbp]
\centering
\caption{Training Recipe for GLM-OCR across different stages.}
\label{tab:training_recipe}
\resizebox{\textwidth}{!}{
\begin{tabular}{lll}
\toprule
\textbf{Stage} & \textbf{Phase} & \textbf{Data Types} \\
\midrule
Stage 1 & Vision Encoder Training & Image-text pairs, Grounding / Retrieval data \\
\addlinespace
Stage 2.1 & Pretrain & Image-text pairs, Document parsing, Grounding, VQA \\
\addlinespace
Stage 2.2 & Pretrain with MTP & Document parsing, Grounding, VQA \\
\addlinespace
Stage 3 & SFT with MTP & Text / Formula / Table recognition, KIE \\
\addlinespace
Stage 4 & RL & Text / Formula / Table recognition, KIE \\
\bottomrule
\end{tabular}
}
\end{table}

\paragraph{Stage 1: Vision Encoder Training.} We first train the vision encoder using large-scale image-text and grounding data to establish strong visual representation capabilities. In this stage, the model is trained on a dataset scaled up to tens of billions of image-text pairs. The training incorporates a dual objective of MIM and CLIP tasks. Furthermore, we employ knowledge distillation from an in-house ViT with a larger parameter size to further enhance the encoder's feature extraction capability.

\paragraph{Stage 2: Vision-Language Pretraining.} In Stage 2.1, we append GLM-0.5B to the Vision Transformer (ViT) and jointly pretrain the full model on image-text, document parsing, grounding, and VQA data to align multimodal representations. In Stage 2.2, we introduce the Multi-Token Prediction (MTP) objective to adapt the decoder for efficient structured generation.

\paragraph{Stage 3: Supervised Fine-Tuning (SFT).} In this stage, we fine-tune the model on curated OCR datasets covering text recognition, formula transcription, table structure recovery, and key information extraction. The objective is to specialize the model for high-precision structured outputs under real-world document distributions. Multi-Token Prediction remains enabled to ensure consistency between training and inference. The data mixture is balanced to prevent overfitting to any single sub-task and to maintain cross-task generalization.

\paragraph{Stage 4: Reinforcement Learning (RL).}

The final stage applies GRPO~\cite{shao2024deepseekmath} to improve structured output reliability and task-specific accuracy. Training samples are generated via rollout from the SFT model, evaluated automatically, and stratified by difficulty to construct a graded optimization set.

The reward function is task-aware and integrates both accuracy-based metrics and structural validation signals. The design is summarized in Table~\ref{tab:rl_reward}.

\begin{table}[htbp]
\centering
\caption{Reward function design in Stage 4 RL training.}
\label{tab:rl_reward}
\resizebox{\textwidth}{!}{
\begin{tabular}{lll}
\toprule
\textbf{Task} & \textbf{Primary Accuracy Reward} & \textbf{Additional Constraints} \\
\midrule
Text Recognition 
& Normalized Edit Distance 
& Repetition penalty \\

Formula Recognition 
& CDM score 
& Structural validity check \\

Table Recognition 
& TEDS score 
& Tag closure verification, structural parsing \\

Key Information Extraction (KIE) 
& Field-level F1 score 
& JSON parse validation, missing/duplicate field penalty \\

\midrule
\multicolumn{3}{l}{\textbf{Global Regularization:} Repetition ratio penalty, malformed structure penalty}
\\
\bottomrule
\end{tabular}
}
\end{table}

\section{Evaluation}

In this section, we evaluate the performance of GLM-OCR against current state-of-the-art pipeline tools, general Vision-Language Models (VLMs), and specialized OCR VLMs. To provide a comprehensive assessment, the evaluation is divided into two parts: standard Public Benchmarks and custom In-House Benchmarks.

\subsection{Public Benchmarks}

We first evaluate GLM-OCR on widely recognized public datasets encompassing Document Parsing and KIE tasks.

\paragraph{Overall Benchmark Performance.} As shown in Table~\ref{tab:docparse}, GLM-OCR demonstrates superior performance across the majority of standard datasets. In Document Parsing, our model achieves the highest scores on OmniDocBench v1.5 (94.6), OCRBench Text (94.0), UniMERNet (96.5), and TEDS\_TEST (86.0). It remains highly competitive on PubTabNet (85.2), trailing only MinerU 2.5. Furthermore, GLM-OCR establishes a clear SOTA in KIE, outperforming all available open-source competitors on Nanonets-KIE (93.7) and Handwritten-KIE (86.1), and even narrowing the gap with closed-source giants like Gemini-3-Pro.

\begin{table*}[t]
\centering
\caption{Performance comparison of various OCR models. The results of Gemini-3-Pro and GPT-5.2-2025-12-11 are in gray as they are provided for reference only and excluded from the best-score ranking. The best results among the evaluated models are highlighted in bold.}
\label{tab:docparse}:
\resizebox{\textwidth}{!}{
\begin{tabular}{l ccccc cc}
\toprule
\textbf{Dataset} & \textbf{GLM-OCR} & \makecell{\textbf{PaddleOCR}\\\textbf{-VL-1.5}} & \makecell{\textbf{Deepseek}\\\textbf{-OCR2}} & \makecell{\textbf{MinerU}\\\textbf{2.5}} & \textbf{dots.ocr} & \makecell{\textcolor{gray}{\textbf{Gemini-3}}\\\textcolor{gray}{\textbf{-Pro}}} & \makecell{\textcolor{gray}{\textbf{GPT-5.2}}\\\textcolor{gray}{\textbf{-2025-12-11}}} \\
\midrule
\multicolumn{8}{l}{\textit{\textbf{Document Parsing}}} \\
\midrule
OmniDocBench v1.5 & \textbf{94.6} & 94.5 & 91.1 & 90.7 & 88.4 & \textcolor{gray}{90.3} & \textcolor{gray}{85.4} \\
OCRBench (Text)   & \textbf{94.0} & 75.3 & 34.7 & 75.3 & 92.1 & \textcolor{gray}{91.9} & \textcolor{gray}{83.7} \\
UniMERNet         & \textbf{96.5} & 96.1 & 85.8 & 96.4 & 90.0 & \textcolor{gray}{96.4} & \textcolor{gray}{90.5} \\
PubTabNet         & 85.2 & 84.6 & -    & \textbf{88.4} & 71.0 & \textcolor{gray}{91.4} & \textcolor{gray}{84.4} \\
TEDS\_TEST        & \textbf{86.0} & 83.3 & -    & 85.4 & 62.4 & \textcolor{gray}{81.8} & \textcolor{gray}{67.6} \\
\midrule
\multicolumn{8}{l}{\textit{\textbf{Key Information Extraction}}} \\
\midrule
Nanonets-KIE      & \textbf{93.7} & -    & -    & -    & -    & \textcolor{gray}{95.2} & \textcolor{gray}{87.5} \\
Handwritten-KIE   & \textbf{86.1} & -    & -    & -    & -    & \textcolor{gray}{94.5} & \textcolor{gray}{78.2} \\
\bottomrule
\end{tabular}
}
\end{table*}

\paragraph{OmniDocBench v1.5 Analysis.} To better understand the model's document parsing capabilities, we present a granular breakdown of the OmniDocBench v1.5 results in Table~\ref{tab:main_results}. This benchmark compares pipeline tools, general VLMs of varying sizes, and specialized VLMs.

Remarkably, despite possessing only 0.9B parameters, GLM-OCR achieves the highest Overall score (94.62), outperforming not only direct specialized competitors like PaddleOCR-VL-1.5 (94.50) and MinerU2.5 (90.67) but also massive general VLMs such as Qwen3-VL-235B (89.15) and Gemini-3 Pro (90.33).

A breakdown of sub-metrics indicates strong performance in table structure recovery. It achieves the absolute best scores in table recognition, scoring 93.96 on Table$_{TEDS}$ and 96.39 on Table$_{TEDS-S}$. While PaddleOCR-VL-1.5 slightly edges out GLM-OCR in Text$_{Edit}$ (0.035 vs 0.040) and Formula$_{CDM}$ (94.21 vs 93.90), GLM-OCR's exceptional table parsing capabilities secure its position as the top-performing model overall, proving that specialized, parameter-efficient architectures can rival or surpass scale-heavy models in complex document parsing.

\begin{table*}[t]
\centering
\caption{Performance comparison with pipeline tools, general VLMs and specialized VLMs on OmniDocBench v1.5.}
\resizebox{\textwidth}{!}{
\begin{tabular}{l l c c c c c c c}
\toprule
\textbf{Model Type} & \textbf{Methods} & \textbf{Params} 
& \makecell{\textbf{Overall}$\uparrow$} 
& \makecell{\textbf{Text}$\downarrow$ \\ \textbf{Edit}}
& \makecell{\textbf{Formula}$\uparrow$ \\ \textbf{CDM}}
& \makecell{\textbf{Table}$\uparrow$ \\ \textbf{TEDS}} 
& \makecell{\textbf{Table}$\uparrow$ \\ \textbf{TEDS-S}} 
& \makecell{\textbf{Reading Order}$\downarrow$ \\ \textbf{Edit}} \\
\midrule

\multirow{3}{*}{Pipeline Tools}
& Marker-1.8.2~\cite{Marker-1.8.2} & - & 71.30 & 0.206 & 76.66 & 57.88 & 71.17 & 0.250 \\
& Mineru2-pipeline~\cite{Mineru2-pipeline} & - & 75.51 & 0.209 & 76.55 & 70.90 & 79.11 & 0.225 \\
& PP-StructureV3~\cite{PaddleOCR} & - & 86.73 & 0.073 & 85.79 & 81.68 & 89.48 & 0.073 \\
\midrule

\multirow{8}{*}{General VLMs}
& GPT-4o~\cite{GPT-4o} & - & 75.02 & 0.217 & 79.70 & 67.07 & 76.09 & 0.148 \\
& InternVL3-76B~\cite{zhu2025internvl3} & 76B & 80.33 & 0.131 & 83.42 & 70.64 & 77.74 & 0.113 \\
& InternVL3.5-241B~\cite{wang2025internvl35} & 241B & 82.67 & 0.142 & 87.23 & 75.00 & 81.28 & 0.125 \\
& GPT-5.2~\cite{gpt5_2} & - & 85.50 & 0.123 & 86.11 & 82.66 & 87.35 & 0.099 \\
& Qwen2.5-VL-72B~\cite{bai2025qwen2} & 72B & 87.02 & 0.094 & 88.27 & 82.15 & 86.22 & 0.102 \\
& Gemini-2.5 Pro~\cite{gemini25} & - & 88.03 & 0.075 & 85.82 & 85.71 & 90.29 & 0.097 \\
& Qwen3-VL~\cite{Qwen3-VL} & 235B & 89.15 & 0.069 & 88.14 & 86.21 & 90.55 & 0.068 \\
& Gemini-3 Pro~\cite{gemini30} & - & 90.33 & 0.065 & 89.18 & 88.28 & 90.29 & 0.071 \\
\midrule

\multirow{15}{*}{Specialized VLMs}
& Dolphin~\cite{feng2025dolphin} & 0.3B & 74.67 & 0.125 & 67.85 & 68.70 & 77.77 & 0.124 \\
& OCRFlux-3B~\cite{OCRFlux2025} & 3B & 74.82 & 0.193 & 68.03 & 75.75 & 80.23 & 0.202 \\
& Mistral OCR~\cite{mistral} & - & 78.83 & 0.164 & 82.84 & 70.03 & 78.04 & 0.144 \\
& POINTS-Reader~\cite{liu2025points} & 3B & 80.98 & 0.134 & 79.20 & 77.13 & 81.66 & 0.145 \\
& olmOCR-7B~\cite{poznanski2025olmocr} & 7B & 81.79 & 0.096 & 86.04 & 68.92 & 74.77 & 0.121 \\
& Dolphin-1.5~\cite{feng2025dolphin} & 0.3B & 83.21 & 0.092 & 80.78 & 78.06 & 84.10 & 0.080 \\
& MinerU2-VLM~\cite{Mineru2-pipeline} & 0.9B & 85.56 & 0.078 & 80.95 & 83.54 & 87.66 & 0.086 \\
& Nanonets-OCR-s~\cite{Nanonets-OCR-S} & 3B & 85.59 & 0.093 & 85.90 & 80.14 & 85.57 & 0.108 \\
& MonkeyOCR-pro-1.2B~\cite{li2025monkeyocr} & 1.9B & 86.96 & 0.084 & 85.02 & 84.24 & 89.02 & 0.130 \\
& Deepseek-OCR~\cite{wei2025deepseek} & 3B & 87.01 & 0.073 & 83.37 & 84.97 & 88.80 & 0.086 \\
& MonkeyOCR-3B~\cite{li2025monkeyocr} & 3.7B & 87.13 & 0.075 & 87.45 & 81.39 & 85.92 & 0.129 \\
& dots.ocr~\cite{dotsocr} & 3B & 88.41 & 0.048 & 83.22 & 86.78 & 90.62 & 0.053 \\
& MonkeyOCR-pro-3B~\cite{li2025monkeyocr} & 3.7B & 88.85 & 0.075 & 87.25 & 86.78 & 90.63 & 0.128 \\
& MinerU2.5~\cite{niu2025mineru2} & 1.2B & 90.67 & 0.047 & 88.46 & 88.22 & 92.38 & 0.044 \\
& PaddleOCR-VL~\cite{PaddleOCR-VL} & 0.9B & 92.86 & 0.035 & 91.22 & 90.89 & 94.76 & 0.043 \\
& PaddleOCR-VL-1.5~\cite{PaddleOCR-VL-1.5} & 0.9B & 94.50 & \textbf{0.035} & \textbf{94.21} & 92.76 & 95.79 & \textbf{0.042} \\
\midrule
& \textbf{GLM-OCR} & 0.9B & \textbf{94.62} & 0.040 & 93.90 & \textbf{93.96} & \textbf{96.39} & 0.044 \\
\bottomrule
\end{tabular}}
\label{tab:main_results}
\end{table*}

\subsection{In-House Benchmarks}

To assess the robustness of GLM-OCR in highly complex, real-world industrial scenarios, we conducted evaluations on a custom suite of in-house benchmarks. These tasks include Code Document parsing, Real-world Table extraction, Handwritten Text recognition, Multilingual Text processing, Seal Recognition, and Receipt KIE.

\begin{table*}[t]
\centering
\caption{Performance comparison on custom real-world scenarios. The results of Gemini-3-Pro and GPT-5.2-2025-12-11 are in gray as they are provided for reference only and excluded from the best-score ranking. The best results among the evaluated models are highlighted in bold.}
\label{tab:self_eval}
\resizebox{\textwidth}{!}{
\begin{tabular}{l ccccc cc}
\toprule
\textbf{Task} & \textbf{GLM-OCR} & \makecell{\textbf{PaddleOCR}\\\textbf{-VL-1.5}} & \makecell{\textbf{Deepseek}\\\textbf{-OCR2}} & \makecell{\textbf{MinerU}\\\textbf{2.5}} & \textbf{dots.ocr} & \makecell{\textcolor{gray}{\textbf{Gemini-3}}\\\textcolor{gray}{\textbf{-Pro}}} & \makecell{\textcolor{gray}{\textbf{GPT-5.2}}\\\textcolor{gray}{\textbf{-2025-12-11}}} \\
\midrule
Code Document     & \textbf{84.7} & 75.8 & 82.1 & 82.9 & 80.8 & \textcolor{gray}{86.9} & \textcolor{gray}{84.4} \\
Real-world Table  & \textbf{91.5} & 86.1 & -    & 70.8 & 81.8 & \textcolor{gray}{90.6} & \textcolor{gray}{86.7} \\
Handwritten Text  & 87.0 & \textbf{87.4} & 73.8 & 54.2 & 71.7 & \textcolor{gray}{90.0} & \textcolor{gray}{78.0} \\
Multilingual Text & \textbf{69.3} & 54.8 & 56.1 & 27.8 & 65.1 & \textcolor{gray}{86.2} & \textcolor{gray}{70.1} \\
Seal Recognition  & \textbf{90.5} & 42.2 & 40.4 & -    & 63.0 & \textcolor{gray}{91.3} & \textcolor{gray}{58.8} \\
Receipt KIE       & \textbf{94.5} & -    & -    & -    & -    & \textcolor{gray}{97.3} & \textcolor{gray}{83.5} \\
\bottomrule
\end{tabular}
}
\end{table*}

As detailed in Table~\ref{tab:self_eval}, GLM-OCR achieves the highest score in five out of six evaluated categories among the compared open-weight models. Most notably, GLM-OCR demonstrates a notable margin in challenging, niche domains:

\begin{itemize}
    \item \textbf{Seal Recognition:} GLM-OCR achieves an exceptional score of 90.5, outperforming the next best open-weight model (dots.ocr at 63.0) by a massive margin and performing competitively with Gemini-3-Pro (91.3).
    \item \textbf{Multilingual Text:} The model excels in diverse linguistic contexts, scoring 69.3 compared to PaddleOCR-VL-1.5's 54.8.
    \item \textbf{Complex Formatting:} It leads in Code Document parsing (84.7) and Real-world Table extraction (91.5), proving its utility in highly structured and noisy environments.
\end{itemize}

While PaddleOCR-VL-1.5 holds a marginal lead in Handwritten Text (87.4 vs. GLM-OCR's 87.0), GLM-OCR remains highly effective. Crucially, in practical application tasks like Receipt KIE, GLM-OCR (94.5) easily surpasses proprietary models like GPT-5.2 (83.5). These in-house results validate that GLM-OCR is not merely optimizing for academic datasets but is highly capable of generalizing to the noisy, variable conditions of real-world OCR deployments.

\section{Inference and Deployment}
\label{sec:inference_deployment}

\subsection{Local Deployment and SDK Integration}
With a compact parameter scale of 0.9B, GLM-OCR is highly optimized for localized inference and resource-constrained environments. The model supports efficient deployment across mainstream frameworks, including vLLM, SGLang, and Ollama. To facilitate seamless integration, a comprehensive SDK is provided for end-to-end document parsing workflows~\footnote{https://github.com/zai-org/GLM-OCR}.

To assess operational efficiency, we conducted a comparative throughput analysis of various OCR pipelines. Under identical hardware configurations and testing conditions (single replica, single concurrency), we evaluated the parsing and Markdown-export speeds for both image and PDF inputs. As demonstrated in Table~\ref{tab:ocr_throughput}, GLM-OCR outperforms comparable methods, achieving a throughput of 1.86 pages/second for PDF documents and 0.67 images/second for standalone image files.

\begin{table}[htbp]
    \centering
    \caption{Throughput comparison of different OCR models for Image and PDF inputs.}
    \label{tab:ocr_throughput}
    \begin{tabular}{lcc}
        \toprule
        \multirow{2}{*}{\textbf{Model}} & \textbf{Image Input} & \textbf{PDF Input} \\
        & \textbf{(pages / s)} & \textbf{(pages / s)} \\
        \midrule
        GLM-OCR & \textbf{0.67} & \textbf{1.86} \\
        PaddleOCR-VL-1.5 & 0.39 & 1.22 \\
        Deepseek-OCR2 & 0.32 & $-$ \\
        MinerU2.5 & 0.18 & 0.48 \\
        dots.ocr & 0.10 & $-$ \\
        \bottomrule
    \end{tabular}
\end{table}

\subsection{Model-as-a-Service (MaaS) API}
For cloud-based deployments, GLM-OCR is accessible via a MaaS API
~\footnote{https://docs.bigmodel.cn/cn/guide/models/vlm/glm-ocr}. The service employs a highly cost-effective, unified pricing model for both input and output tokens, set at 0.2 RMB per million tokens. Under this pricing structure, an expenditure of 1 RMB is sufficient to process approximately 2,000 A4-sized scanned images or 200 simple-layout PDFs (10 pages each). This represents a significant reduction in operational overhead, decreasing processing costs to approximately one-tenth of those associated with traditional OCR solutions.

\subsection{Fine-Tuning Capabilities}
In scenarios where specific domain adaptation or enhanced task performance is required, GLM-OCR supports direct fine-tuning utilizing the LLaMA-Factory framework. Comprehensive tutorials and configuration guidelines for the fine-tuning process are available in the official repository: \url{https://github.com/zai-org/GLM-OCR/blob/main/examples/finetune/README.md}.

\section{Intended Use Cases}

\subsection{Overview}

GLM-OCR is designed to support both high-level document understanding workflows and lightweight optical character recognition (OCR) tasks. Depending on application requirements, users may (1) integrate the GLM-OCR SDK for complex document parsing pipelines, or (2) directly invoke the base model for focused recognition and structured information extraction tasks. This section describes the two primary usage paradigms and their representative application scenarios.

\subsection{Document Parsing with GLM-OCR SDK}

The GLM-OCR SDK~\footnote{https://github.com/zai-org/GLM-OCR} provides a comprehensive interface for performing document parsing, including layout-aware parsing, multimodal recognition, and structured output generation. It is intended for enterprise-grade or production-level workflows where documents may contain heterogeneous content such as paragraphs, tables, mathematical expressions, and key-value pairs.

\begin{figure}[!h]
\centering
\includegraphics[width=\linewidth]{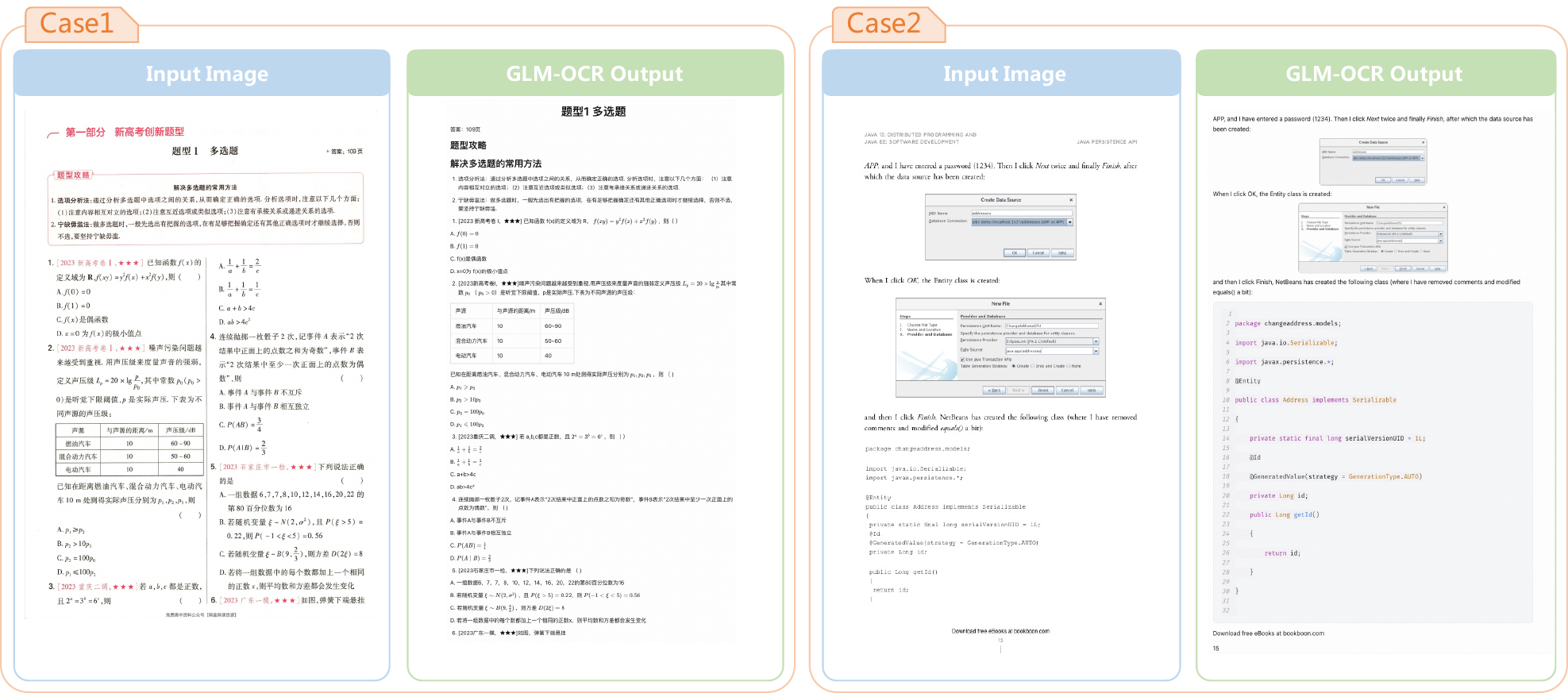}
\caption{Example of GLM-OCR SDK for complex document parsing.}
\label{fig:use_case_sdk}
\end{figure}

The output is generated in a structured markdown format, preserving logical document hierarchy and structural relationships. The example demonstrates that the SDK supports end-to-end parsing of heterogeneous documents while maintaining structural fidelity and semantic coherence.

\subsection{Lightweight OCR and Information Extraction with the Base Model}

In addition to the SDK, GLM-OCR can be directly used as a standalone model for lightweight OCR and targeted extraction tasks. This mode is appropriate for scenarios requiring lower integration overhead, flexible prompting, or rapid prototyping.

All tasks in this mode are controlled via explicit prompt instructions. Below we describe four primary task categories.

\subsubsection{Text Recognition}

\textbf{Prompt:}
\begin{verbatim}
Text Recognition:
\end{verbatim}

This mode is used for general printed or handwritten text transcription. The model outputs plain text corresponding to the visible textual content in the input image.

\paragraph{Example Scenario.}

\begin{figure}[!h]
\centering
\includegraphics[width=\linewidth]{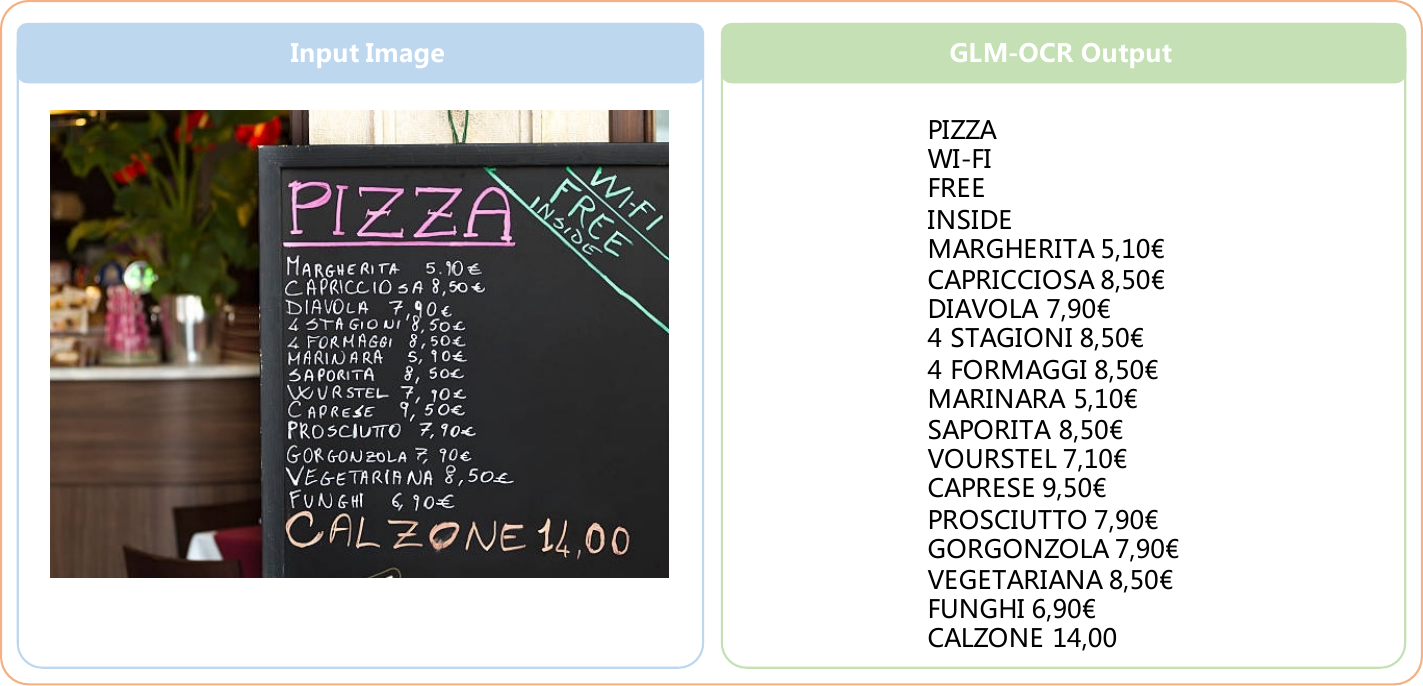}
\caption{Example of GLM-OCR Text Recognition.}
\label{fig:use_case_text}
\end{figure}

Figure~\ref{fig:use_case_text} illustrates a real-world text recognition scenario using a restaurant menu board containing multilingual content (e.g., Italian dish names and English phrases) and price annotations with special characters (e.g., the euro symbol). The input image exhibits typical challenges such as handwritten-style fonts, varying character sizes, non-uniform spacing, perspective distortion, and background clutter.

Despite these complexities, the model accurately transcribes the textual content while preserving the original line structure and semantic grouping. In particular, it correctly reconstructs line breaks, capitalization, punctuation, numerical values, and currency symbols (e.g., “5,10€”, “14,00”), demonstrating robustness to layout variations and moderate visual noise. This example highlights the model’s ability to perform reliable optical character recognition in unconstrained, real-world environments.

\subsubsection{Table Recognition}

\textbf{Prompt:}
\begin{verbatim}
Table Recognition:
\end{verbatim}

This task focuses on recovering the structural representation of tabular data. The output may be formatted as Markdown tables or structured text that preserves row and column alignment.

\paragraph{Example Scenario.}

\begin{figure}[!h]
\centering
\includegraphics[width=\linewidth]{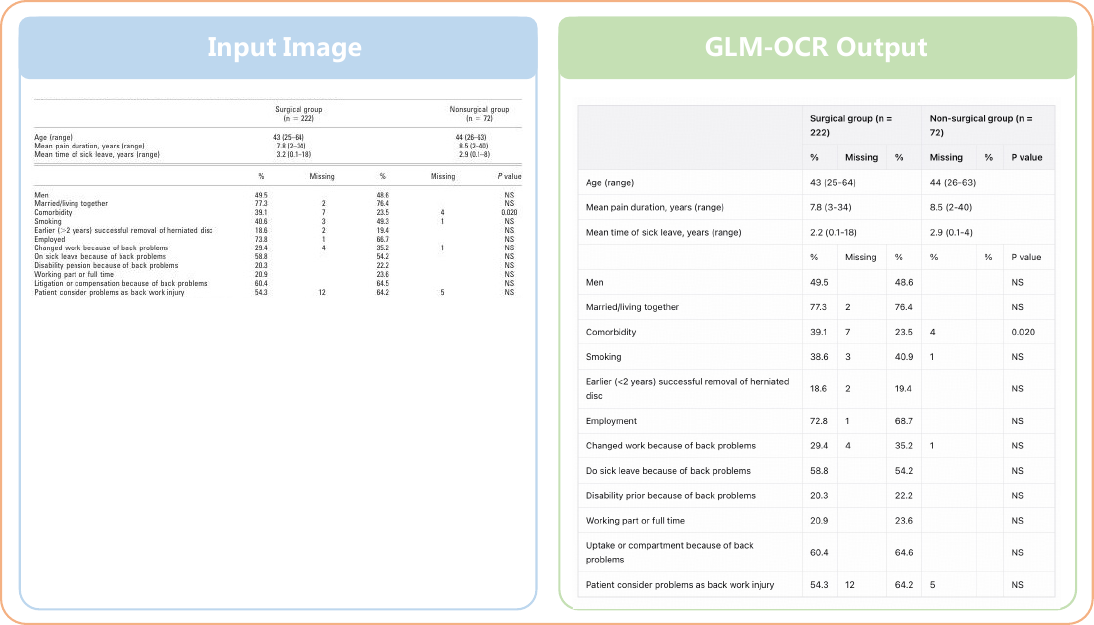}
\caption{Example of GLM-OCR Table Recognition.}
\label{fig:use_case_table}
\end{figure}

Figure~\ref{fig:use_case_table} presents a clinical summary table containing hierarchical column headers, merged cells, percentage values, missing-value indicators, and statistical annotations (e.g., p-values). The input image exhibits typical document-analysis challenges, including low contrast, dense numerical content, multi-level header grouping, and complex row–column alignment.

GLM-OCR accurately reconstructs the logical table structure by identifying column groups (e.g., surgical vs.\ non-surgical cohorts), preserving header hierarchies, and correctly aligning numerical entries with their corresponding attributes. In addition to faithfully transcribing cell content, the model maintains structural consistency such as row ordering, column correspondence, and percentage–value pairing. This structured reconstruction enables direct conversion into machine-readable formats (e.g., CSV or spreadsheet tables), thereby facilitating downstream statistical analysis, data validation, and automated reporting workflows.

\subsubsection{Formula Recognition}

\textbf{Prompt:}
\begin{verbatim}
Formula Recognition:
\end{verbatim}

This mode is intended for recognizing mathematical expressions and converting them into structured formats (e.g., LaTeX).

\paragraph{Example Scenario.}

\begin{figure}[!h]
\centering
\includegraphics[width=\linewidth]{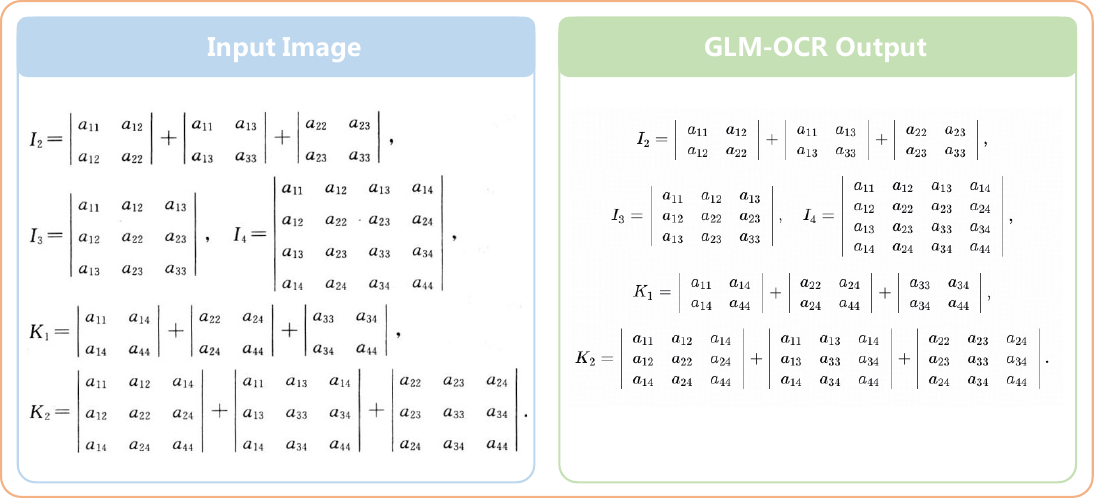}
\caption{Example of GLM-OCR Formula Recognition.}
\label{fig:use_case_formula}
\end{figure}

Given an image containing inline and display equations from a scientific manuscript, the model transcribes the formulas into syntactically valid LaTeX expressions. The output preserves operators, superscripts, subscripts, and fraction structures, enabling direct reuse in academic or technical documentation.

As depicted in Figure \ref{fig:use_case_formula}, given an image containing dense mathematical equations, GLM-OCR accurately transcribes the visual content into syntactically valid LaTeX expressions. The model demonstrates high fidelity in parsing complex two-dimensional spatial layouts, successfully preserving intricate structural elements such as matrices, determinants, and multi-level subscripts. This precise reconstruction eliminates the need for manual correction, enabling direct and seamless reuse in academic and technical documentation.

\subsubsection{Key Information Extraction}

For structured information extraction tasks, the prompt must explicitly specify a strict JSON schema. The model is expected to generate output that conforms to the provided format.



\paragraph{Example Scenario.}

\begin{figure}[!h]
\centering
\includegraphics[width=\linewidth]{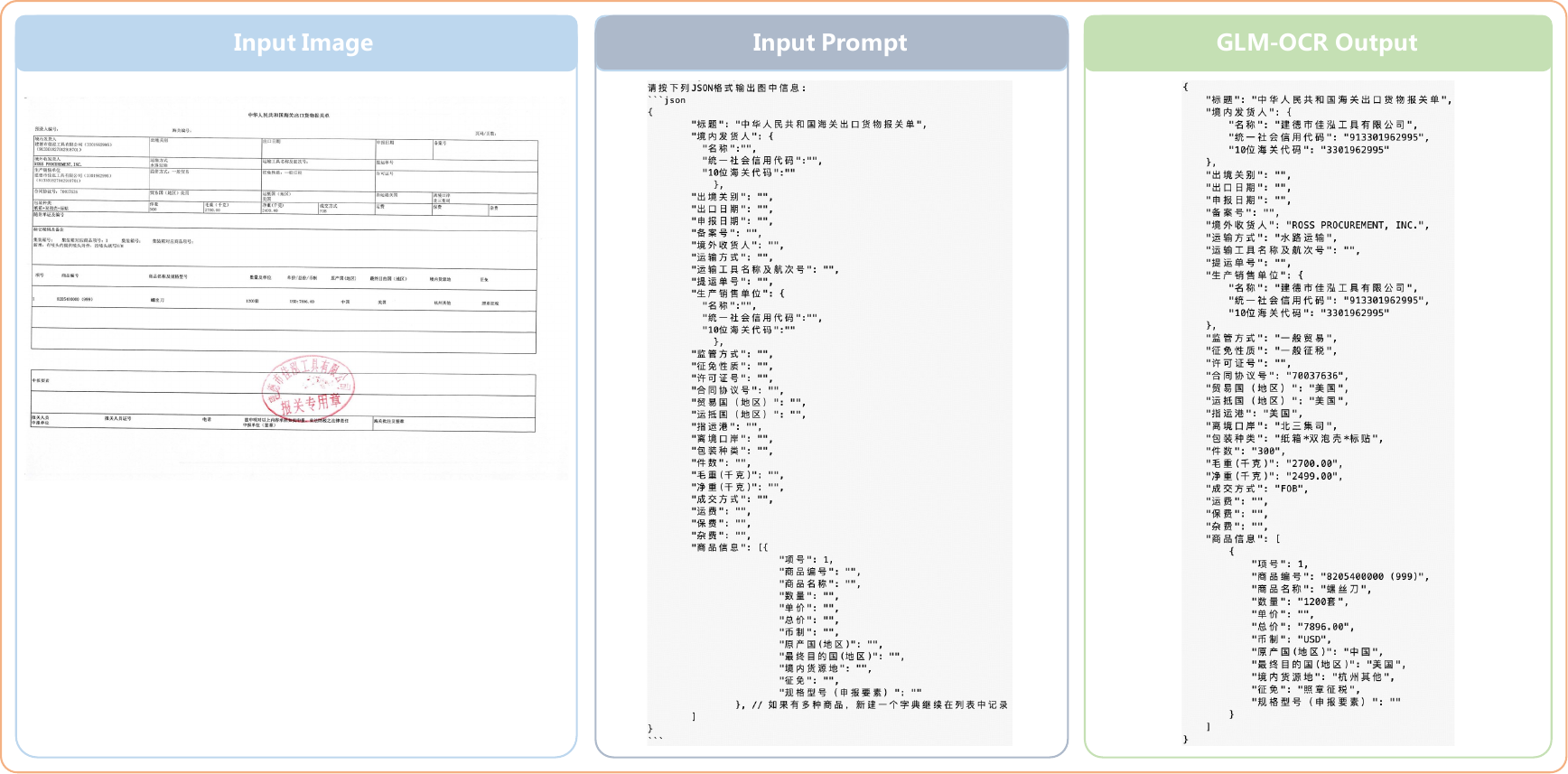}
\caption{Example of GLM-OCR Key Information Extraction.}
\label{fig:use_case_kie}
\end{figure}

In a representative example involving a dense customs declaration form, the model successfully extracts structured fields based on a detailed input prompt. It accurately populates complex, nested JSON entities—such as shipper details, unified social credit codes, and itemized goods information—directly from the visual layout. The output strictly adheres to the user-provided JSON schema, eliminating hallucinated keys and facilitating seamless integration into automated processing pipelines or structured databases.

\subsection{Summary of Usage Paradigms}

The two usage modes address complementary needs:

\begin{itemize}
    \item \textbf{GLM-OCR SDK} is designed for comprehensive, layout-aware, multi-element document parsing in production environments.
    \item \textbf{Base model prompting} enables lightweight, flexible OCR and targeted information extraction for modular or task-specific applications.
\end{itemize}

Together, these paradigms allow GLM-OCR to support a broad spectrum of document understanding workflows, ranging from rapid prototyping to enterprise-scale deployment.

\section{Limitations}

Although GLM-OCR demonstrates competitive performance across diverse benchmarks and practical scenarios, several limitations remain.

\subsection{Two-Stage Architectural Constraints}

The current two-stage pipeline, consisting of layout analysis followed by region-level recognition, may introduce error propagation. In cases of inaccurate layout detection, downstream recognition performance may degrade. Additionally, complex layouts involving cross-page dependencies or irregular multi-column structures may lead to imperfect reading order reconstruction.

\subsection{Data Coverage Limitations}

Model performance is influenced by the distribution and diversity of training data. Degradation may occur in scenarios involving:

\begin{itemize}
    \item Extremely low-resolution or heavily distorted documents,
    \item Highly complex mathematical expressions,
    \item Dense or irregular tabular structures,
    \item Underrepresented languages in the training corpus.
\end{itemize}

\subsection{Structured Output Variability}

As a generative model, GLM-OCR may exhibit minor stochastic variation in formatting behaviors, particularly in line breaks and whitespace handling. Although reinforcement learning and structural supervision mitigate this effect, strict formatting guarantees cannot be fully ensured.

\subsection{Key Information Extraction}

While the model supports prompt-based KIE, extraction accuracy depends on prompt specification and schema clarity. In complex forms with implicit or ambiguous field boundaries, incomplete or redundant outputs may occur.

These limitations represent areas for continued research and system refinement.

\section{Conclusion}

This report introduces GLM-OCR as a practical solution for structured document understanding under real-world system constraints. Instead of relying on large model scaling, the design prioritizes controllable latency, memory efficiency, and structured output reliability. Through the combination of layout-aware preprocessing and multi-token decoding, the system improves throughput and stability while maintaining competitive recognition accuracy across diverse document types. The results indicate that careful alignment between model architecture, decoding strategy, and task structure can yield substantial efficiency gains without increasing parameter scale.

From an engineering perspective, GLM-OCR demonstrates that document intelligence systems benefit from modular pipelines, efficient generation mechanisms, and deployment-oriented optimization. The model supports local inference, cloud-based serving, and domain-specific fine-tuning, enabling integration into heterogeneous production environments. Future development will focus on improving robustness under extreme layout complexity, enhancing multilingual coverage, and strengthening structured output consistency to further reduce downstream integration costs.

\bibliography{iclr2026_conference}

@inproceedings{Tesseract,
  title={An overview of the Tesseract OCR engine},
  author={Smith, Ray},
  booktitle={Ninth international conference on document analysis and recognition (ICDAR 2007)},
  volume={2},
  pages={629--633},
  year={2007},
  organization={IEEE}
}

@misc{PaddleOCR,
      title={PaddleOCR 3.0 Technical Report}, 
      author={Cheng Cui and Ting Sun and Manhui Lin and Tingquan Gao and Yubo Zhang and Jiaxuan Liu and Xueqing Wang and Zelun Zhang and Changda Zhou and Hongen Liu and Yue Zhang and Wenyu Lv and Kui Huang and Yichao Zhang and Jing Zhang and Jun Zhang and Yi Liu and Dianhai Yu and Yanjun Ma},
      year={2025},
      eprint={2507.05595},
      archivePrefix={arXiv},
      primaryClass={cs.CV},
      url={https://arxiv.org/abs/2507.05595}, 
}

@misc{EasyOCR,
  title = {EasyOCR},
  author = {JaidedAI},
  year = {2020},
  howpublished = {\url{https://github.com/JaidedAI/EasyOCR}},
}

@article{Qwen3-VL,
      title={Qwen3-VL Technical Report}, 
      author={Shuai Bai and Yuxuan Cai and Ruizhe Chen and Keqin Chen and Xionghui Chen and Zesen Cheng and Lianghao Deng and Wei Ding and Chang Gao and Chunjiang Ge and Wenbin Ge and Zhifang Guo and Qidong Huang and Jie Huang and Fei Huang and Binyuan Hui and Shutong Jiang and Zhaohai Li and Mingsheng Li and Mei Li and Kaixin Li and Zicheng Lin and Junyang Lin and Xuejing Liu and Jiawei Liu and Chenglong Liu and Yang Liu and Dayiheng Liu and Shixuan Liu and Dunjie Lu and Ruilin Luo and Chenxu Lv and Rui Men and Lingchen Meng and Xuancheng Ren and Xingzhang Ren and Sibo Song and Yuchong Sun and Jun Tang and Jianhong Tu and Jianqiang Wan and Peng Wang and Pengfei Wang and Qiuyue Wang and Yuxuan Wang and Tianbao Xie and Yiheng Xu and Haiyang Xu and Jin Xu and Zhibo Yang and Mingkun Yang and Jianxin Yang and An Yang and Bowen Yu and Fei Zhang and Hang Zhang and Xi Zhang and Bo Zheng and Humen Zhong and Jingren Zhou and Fan Zhou and Jing Zhou and Yuanzhi Zhu and Ke Zhu},
	  journal={arXiv preprint arXiv:2511.21631},
      year={2025}
}

@misc{glm-4.5v,
      title={GLM-4.5V and GLM-4.1V-Thinking: Towards Versatile Multimodal Reasoning with Scalable Reinforcement Learning},
      author={V Team and Wenyi Hong and Wenmeng Yu and Xiaotao Gu and Guo Wang and Guobing Gan and Haomiao Tang and Jiale Cheng and Ji Qi and Junhui Ji and Lihang Pan and Shuaiqi Duan and Weihan Wang and Yan Wang and Yean Cheng and Zehai He and Zhe Su and Zhen Yang and Ziyang Pan and Aohan Zeng and Baoxu Wang and Bin Chen and Boyan Shi and Changyu Pang and Chenhui Zhang and Da Yin and Fan Yang and Guoqing Chen and Jiazheng Xu and Jiale Zhu and Jiali Chen and Jing Chen and Jinhao Chen and Jinghao Lin and Jinjiang Wang and Junjie Chen and Leqi Lei and Letian Gong and Leyi Pan and Mingdao Liu and Mingde Xu and Mingzhi Zhang and Qinkai Zheng and Sheng Yang and Shi Zhong and Shiyu Huang and Shuyuan Zhao and Siyan Xue and Shangqin Tu and Shengbiao Meng and Tianshu Zhang and Tianwei Luo and Tianxiang Hao and Tianyu Tong and Wenkai Li and Wei Jia and Xiao Liu and Xiaohan Zhang and Xin Lyu and Xinyue Fan and Xuancheng Huang and Yanling Wang and Yadong Xue and Yanfeng Wang and Yanzi Wang and Yifan An and Yifan Du and Yiming Shi and Yiheng Huang and Yilin Niu and Yuan Wang and Yuanchang Yue and Yuchen Li and Yutao Zhang and Yuting Wang and Yu Wang and Yuxuan Zhang and Zhao Xue and Zhenyu Hou and Zhengxiao Du and Zihan Wang and Peng Zhang and Debing Liu and Bin Xu and Juanzi Li and Minlie Huang and Yuxiao Dong and Jie Tang},
      year={2025},
      eprint={2507.01006},
      archivePrefix={arXiv},
      primaryClass={cs.CV},
      url={https://arxiv.org/abs/2507.01006},
}

@article{seed1_5vl,
  title={Seed1.5-VL Technical Report},
  author={ByteDance Seed Team},
  journal={arXiv preprint arXiv:2505.07062},
  year={2025}
}

@misc{glm2024chatglm,
      title={ChatGLM: A Family of Large Language Models from GLM-130B to GLM-4 All Tools},
      author={Team GLM and Aohan Zeng and Bin Xu and Bowen Wang and Chenhui Zhang and Da Yin and Diego Rojas and Guanyu Feng and Hanlin Zhao and Hanyu Lai and Hao Yu and Hongning Wang and Jiadai Sun and Jiajie Zhang and Jiale Cheng and Jiayi Gui and Jie Tang and Jing Zhang and Juanzi Li and Lei Zhao and Lindong Wu and Lucen Zhong and Mingdao Liu and Minlie Huang and Peng Zhang and Qinkai Zheng and Rui Lu and Shuaiqi Duan and Shudan Zhang and Shulin Cao and Shuxun Yang and Weng Lam Tam and Wenyi Zhao and Xiao Liu and Xiao Xia and Xiaohan Zhang and Xiaotao Gu and Xin Lv and Xinghan Liu and Xinyi Liu and Xinyue Yang and Xixuan Song and Xunkai Zhang and Yifan An and Yifan Xu and Yilin Niu and Yuantao Yang and Yueyan Li and Yushi Bai and Yuxiao Dong and Zehan Qi and Zhaoyu Wang and Zhen Yang and Zhengxiao Du and Zhenyu Hou and Zihan Wang},
      year={2024},
      eprint={2406.12793},
      archivePrefix={arXiv},
      primaryClass={id='cs.CL' full_name='Computation and Language' is_active=True alt_name='cmp-lg' in_archive='cs' is_general=False description='Covers natural language processing. Roughly includes material in ACM Subject Class I.2.7. Note that work on artificial languages (programming languages, logics, formal systems) that does not explicitly address natural-language issues broadly construed (natural-language processing, computational linguistics, speech, text retrieval, etc.) is not appropriate for this area.'}
}

@article{deepseekv3,
  title={Deepseek-v3 technical report},
  author={Liu, Aixin and Feng, Bei and Xue, Bing and Wang, Bingxuan and Wu, Bochao and Lu, Chengda and Zhao, Chenggang and Deng, Chengqi and Zhang, Chenyu and Ruan, Chong and others},
  journal={arXiv preprint arXiv:2412.19437},
  year={2024}
}

@article{glm5,
  title={GLM-5: from Vibe Coding to Agentic Engineering},
  author={Zeng, Aohan and Lv, Xin and Hou, Zhenyu and Du, Zhengxiao and Zheng, Qinkai and Chen, Bin and Yin, Da and Ge, Chendi and Xie, Chengxing and Wang, Cunxiang and others},
  journal={arXiv preprint arXiv:2602.15763},
  year={2026}
}

@misc{PaddleOCR-VL,
      title={PaddleOCR-VL: Boosting Multilingual Document Parsing via a 0.9B Ultra-Compact Vision-Language Model}, 
      author={Cheng Cui and Ting Sun and Suyin Liang and Tingquan Gao and Zelun Zhang and Jiaxuan Liu and Xueqing Wang and Changda Zhou and Hongen Liu and Manhui Lin and Yue Zhang and Yubo Zhang and Handong Zheng and Jing Zhang and Jun Zhang and Yi Liu and Dianhai Yu and Yanjun Ma},
      year={2025},
      eprint={2510.14528},
      archivePrefix={arXiv},
      primaryClass={cs.CV},
      url={https://arxiv.org/abs/2510.14528}, 
}

@misc{PaddleOCR-VL-1.5,
      title={PaddleOCR-VL-1.5: Towards a Multi-Task 0.9B VLM for Robust In-the-Wild Document Parsing}, 
      author={Cheng Cui and Ting Sun and Suyin Liang and Tingquan Gao and Zelun Zhang and Jiaxuan Liu and Xueqing Wang and Changda Zhou and Hongen Liu and Manhui Lin and Yue Zhang and Yubo Zhang and Yi Liu and Dianhai Yu and Yanjun Ma},
      year={2026},
      eprint={2601.21957},
      archivePrefix={arXiv},
      primaryClass={cs.CV},
      url={https://arxiv.org/abs/2601.21957}, 
}

@misc{OmniDocBench,
      title={OmniDocBench: Benchmarking Diverse PDF Document Parsing with Comprehensive Annotations}, 
      author={Linke Ouyang and Yuan Qu and Hongbin Zhou and Jiawei Zhu and Rui Zhang and Qunshu Lin and Bin Wang and Zhiyuan Zhao and Man Jiang and Xiaomeng Zhao and Jin Shi and Fan Wu and Pei Chu and Minghao Liu and Zhenxiang Li and Chao Xu and Bo Zhang and Botian Shi and Zhongying Tu and Conghui He},
      year={2024},
      eprint={2412.07626},
      archivePrefix={arXiv},
      primaryClass={cs.CV},
      url={https://arxiv.org/abs/2412.07626}, 
}

@article{OCRBench,
    title={OCRBench: on the hidden mystery of OCR in large multimodal models},
    volume={67},
    ISSN={1869-1919},
    url={http://dx.doi.org/10.1007/s11432-024-4235-6},
    DOI={10.1007/s11432-024-4235-6},
    number={12},
    journal={Science China Information Sciences},
    publisher={Springer Science and Business Media LLC},
    author={Liu, Yuliang and Li, Zhang and Huang, Mingxin and Yang, Biao and Yu, Wenwen and Li, Chunyuan and Yin, Xu-Cheng and Liu, Cheng-Lin and Jin, Lianwen and Bai, Xiang},
    year={2024},
    month=dec
}

@misc{UniMERNet,
      title={UniMERNet: A Universal Network for Real-World Mathematical Expression Recognition}, 
      author={Bin Wang and Zhuangcheng Gu and Guang Liang and Chao Xu and Bo Zhang and Botian Shi and Conghui He},
      year={2024},
      eprint={2404.15254},
      archivePrefix={arXiv},
      primaryClass={cs.CV},
      url={https://arxiv.org/abs/2404.15254}, 
}

@article{PubTabNet,
  title={Image-based table recognition: data, model, and evaluation},
  author={Zhong, Xu and ShafieiBavani, Elaheh and Yepes, Antonio Jimeno},
  journal={arXiv preprint arXiv:1911.10683},
  year={2019}
}

@misc{IDPLeaderboard,
  title={IDPLeaderboard: A Unified Leaderboard for Intelligent Document Processing Tasks},
  author={Souvik Mandal and Nayancy Gupta and Ashish Talewar and Paras Ahuja and Prathamesh Juvatkar and Gourinath Banda},
  howpublished={https://idp-leaderboard.org},
  year={2025},
}

@inproceedings{vllm,
  title={Efficient Memory Management for Large Language Model Serving with PagedAttention},
  author={Woosuk Kwon and Zhuohan Li and Siyuan Zhuang and Ying Sheng and Lianmin Zheng and Cody Hao Yu and Joseph E. Gonzalez and Hao Zhang and Ion Stoica},
  booktitle={Proceedings of the ACM SIGOPS 29th Symposium on Operating Systems Principles},
  year={2023}
}

@article{shao2024deepseekmath,
  title={Deepseekmath: Pushing the limits of mathematical reasoning in open language models},
  author={Shao, Zhihong and Wang, Peiyi and Zhu, Qihao and Xu, Runxin and Song, Junxiao and Bi, Xiao and Zhang, Haowei and Zhang, Mingchuan and Li, YK and Wu, Y and others},
  journal={arXiv preprint arXiv:2402.03300},
  year={2024}
}

@misc{Marker-1.8.2,
  author       = {Vik Paruchuri},
  title        = {Marker},
  year         = {2025},
  howpublished = {\url{https://github.com/datalab-to/marker}},
  note         = {Accessed: 2025-09-25},
}

@misc{MinerU2-pipeline,
  title={MinerU2.0-2505-0.9B},
  howpublished={\url{https://huggingface.co/opendatalab/MinerU2.0-2505-0.9B}},
  author={{opendatalab}},
  year={2025},
}

@article{GPT-4o,
  title={Gpt-4 technical report},
  author={Achiam, Josh and Adler, Steven and Agarwal, Sandhini and Ahmad, Lama and Akkaya, Ilge and Aleman, Florencia Leoni and Almeida, Diogo and Altenschmidt, Janko and Altman, Sam and Anadkat, Shyamal and others},
  journal={arXiv preprint arXiv:2303.08774},
  year={2023}
}

@article{zhu2025internvl3,
  title={Internvl3: Exploring advanced training and test-time recipes for open-source multimodal models},
  author={Zhu, Jinguo and Wang, Weiyun and Chen, Zhe and Liu, Zhaoyang and Ye, Shenglong and Gu, Lixin and Tian, Hao and Duan, Yuchen and Su, Weijie and Shao, Jie and others},
  journal={arXiv preprint arXiv:2504.10479},
  year={2025}
}

@article{wang2025internvl35,
  title={Internvl3. 5: Advancing open-source multimodal models in versatility, reasoning, and efficiency},
  author={Wang, Weiyun and Gao, Zhangwei and Gu, Lixin and Pu, Hengjun and Cui, Long and Wei, Xingguang and Liu, Zhaoyang and Jing, Linglin and Ye, Shenglong and Shao, Jie and others},
  journal={arXiv preprint arXiv:2508.18265},
  year={2025}
}

@misc{gpt5_2,
    title = {GPT-5.2 System Card},
    url = {https://cdn.openai.com/pdf/3a4153c8-c748-4b71-8e31-aecbde944f8d/oai_5_2_system-card.pdf},
    author = {OpenAI.},
    year = {2025}
}

@article{bai2025qwen2,
  title={Qwen2. 5-vl technical report},
  author={Bai, Shuai and Chen, Keqin and Liu, Xuejing and Wang, Jialin and Ge, Wenbin and Song, Sibo and Dang, Kai and Wang, Peng and Wang, Shijie and Tang, Jun and others},
  journal={arXiv preprint arXiv:2502.13923},
  year={2025}
}

@misc{gemini25,
  author={{Google DeepMind}},
  title={Gemini 2.5},
  howpublished={\url{https://blog.google/technology/google-deepmind/gemini-model-thinking-updates-march-2025/}},
  year={2025}
}

@misc{gemini30,
  author={{Google DeepMind}},
  title={Gemini 3.0},
  howpublished={\url{https://blog.google/products-and-platforms/products/gemini/gemini-3-collection/}},
  year={2025}
}

@article{feng2025dolphin,
  title={Dolphin: Document image parsing via heterogeneous anchor prompting},
  author={Feng, Hao and Wei, Shu and Fei, Xiang and Shi, Wei and Han, Yingdong and Liao, Lei and Lu, Jinghui and Wu, Binghong and Liu, Qi and Lin, Chunhui and others},
  journal={arXiv preprint arXiv:2505.14059},
  year={2025}
}

@article{li2025monkeyocr,
  title={MonkeyOCR: Document Parsing with a Structure-Recognition-Relation Triplet Paradigm},
  author={Li, Zhang and Liu, Yuliang and Liu, Qiang and Ma, Zhiyin and Zhang, Ziyang and Zhang, Shuo and Guo, Zidun and Zhang, Jiarui and Wang, Xinyu and Bai, Xiang},
  journal={arXiv preprint arXiv:2506.05218},
  year={2025}
}

@misc{OCRFlux2025,
  author       = {chatdoc-com},
  title        = {OCRFlux},
  year         = {2025},
  howpublished = {\url{https://github.com/chatdoc-com/OCRFlux}},
  note         = {Accessed:2025-09-25},
}

@misc{mistral,
  author={{Mistral AI Team}},
  title={Mistral-ocr},
  howpublished={\url{https://mistral.ai/news/mistral-ocr?utm_source=ai-bot.cn}},
  year={2025}
}

@article{liu2025points,
  title={POINTS-Reader: Distillation-Free Adaptation of Vision-Language Models for Document Conversion},
  author={Liu, Yuan and Zhao, Zhongyin and Tian, Le and Wang, Haicheng and Ye, Xubing and You, Yangxiu and Yu, Zilin and Wu, Chuhan and Zhou, Xiao and Yu, Yang and others},
  journal={arXiv preprint arXiv:2509.01215},
  year={2025}
}

@article{poznanski2025olmocr,
  title={olmocr: Unlocking trillions of tokens in pdfs with vision language models},
  author={Poznanski, Jake and Borchardt, Jon and Dunkelberger, Jason and Huff, Regan and Lin, Daniel and Rangapur, Aman and Wilhelm, Christopher and Lo, Kyle and Soldaini, Luca},
  journal={arXiv preprint arXiv:2502.18443},
  year={2025}
}

@misc{Nanonets-OCR-S,
  title={Nanonets-OCR-S: A model for transforming documents into structured markdown with intelligent content recognition and semantic tagging},
  author={Souvik Mandal and Ashish Talewar and Paras Ahuja and Prathamesh Juvatkar},
  year={2025},
}

@article{wei2025deepseek,
  title={DeepSeek-OCR: Contexts Optical Compression},
  author={Wei, Haoran and Sun, Yaofeng and Li, Yukun},
  journal={arXiv preprint arXiv:2510.18234},
  year={2025}
}

@misc{dotsocr,
  title={dots.ocr: Multilingual Document Layout Parsing in a Single Vision-Language Model},
  author={{rednote-hilab}},
  year={2025},
}

@article{niu2025mineru2,
  title={MinerU2. 5: A Decoupled Vision-Language Model for Efficient High-Resolution Document Parsing},
  author={Niu, Junbo and Liu, Zheng and Gu, Zhuangcheng and Wang, Bin and Ouyang, Linke and Zhao, Zhiyuan and Chu, Tao and He, Tianyao and Wu, Fan and Zhang, Qintong and others},
  journal={arXiv preprint arXiv:2509.22186},
  year={2025}
}
\bibliographystyle{iclr2026_conference}


\end{document}